\documentclass[10pt, conference, compsocconf]{IEEEtran}
\ifCLASSINFOpdf
\else
\fi

\usepackage{graphicx}
\usepackage{caption}
\usepackage{subfig}
\usepackage[cmintegrals]{newtxmath}
\usepackage[T1]{fontenc}
\usepackage{cite}
\usepackage{hyperref}
\usepackage{color}
\usepackage{xcolor}
\usepackage{hyperref}
\usepackage{graphicx}
\usepackage{caption}
\usepackage{subfig}
\usepackage{flafter}
\usepackage{placeins}
\usepackage{fontenc}
\usepackage{stackengine}

\begin{document}
%
\title{Instance Segmentation based Semantic Matting for Compositing Applications}


\author{\IEEEauthorblockN{Guanqing Hu}
\IEEEauthorblockA{
Centre for Intelligent Machines, McGill University\\
Montreal, Canada\\
bonniehu@cim.mcgill.ca}
\and
\IEEEauthorblockN{James J. Clark}
\IEEEauthorblockA{
Centre for Intelligent Machines, McGill University\\
Montreal, Canada\\
clark@cim.mcgill.ca}
}


%


\twocolumn[{%
\renewcommand\twocolumn[1][]{#1}%
\maketitle
\begin{center}
    \centering
    \captionsetup{hypcap=false}
    \stackunder[2pt]{\includegraphics[height=2.5cm,trim=15 15 0 0,clip]{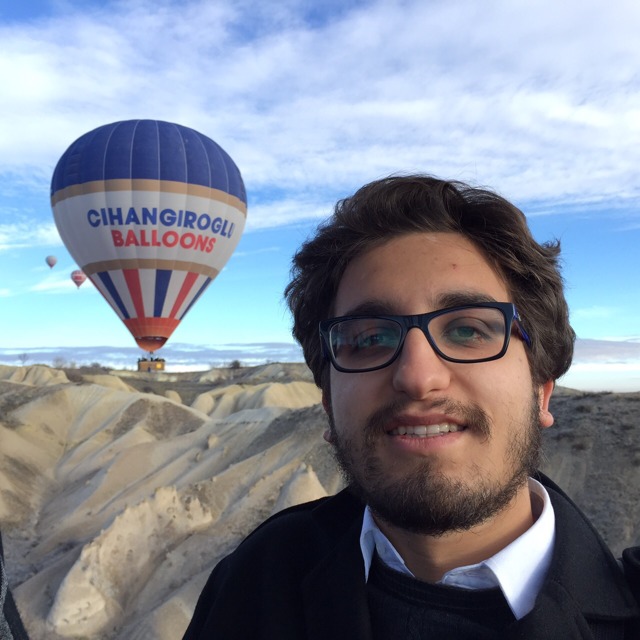}}{\footnotesize Original image}\hspace{4pt}
    \stackunder[2pt]{\includegraphics[height=2.5cm,trim=15 15 0 0,clip]{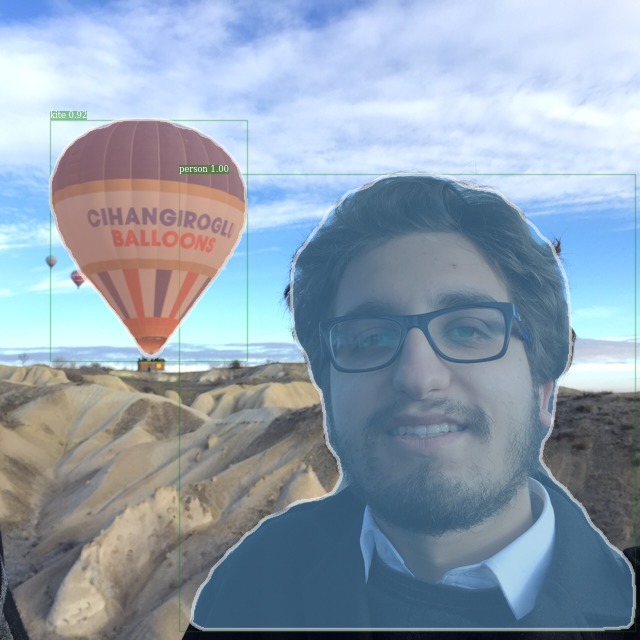}}{\footnotesize Instance segmentation}\hspace{4pt}
   \stackunder[2pt]{\includegraphics[height=2.5cm,trim=15 15 0 0,clip]{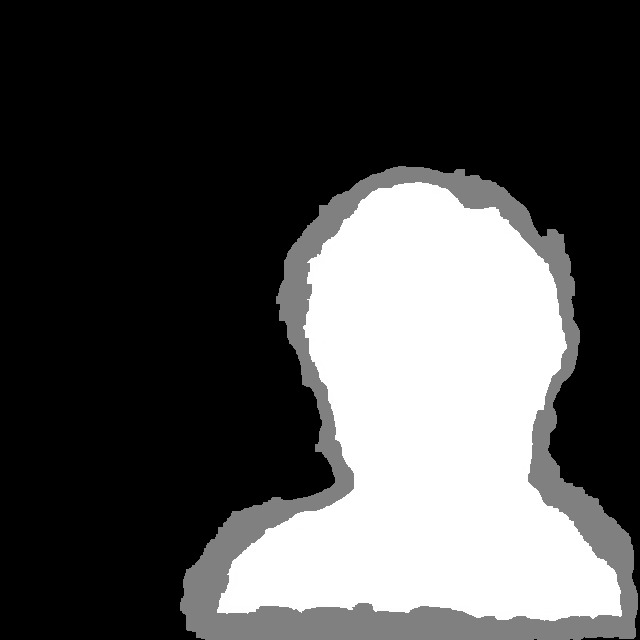}}{\footnotesize Instance 1 trimap}\hspace{4pt}
   \stackunder[2pt]{\includegraphics[height=2.5cm,trim=15 15 0 0,clip]{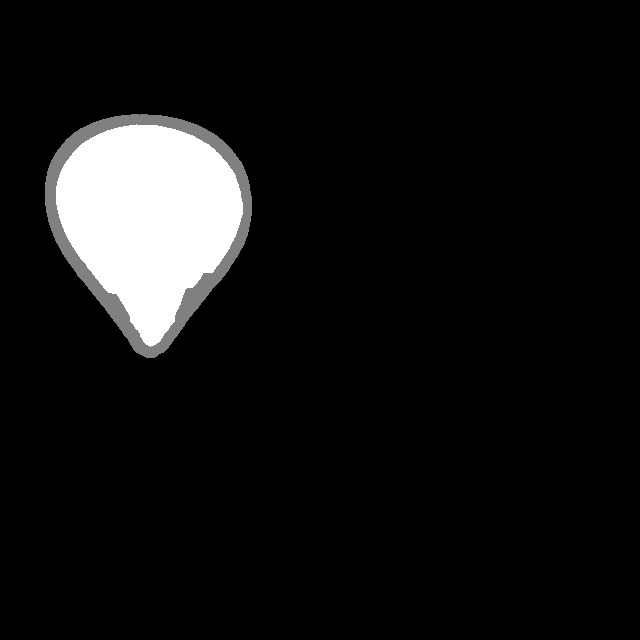}}{\footnotesize Instance 2 trimap}\hspace{4pt}
    \stackunder[2pt]{\includegraphics[height=2.5cm,trim=15 15 0 0,clip]{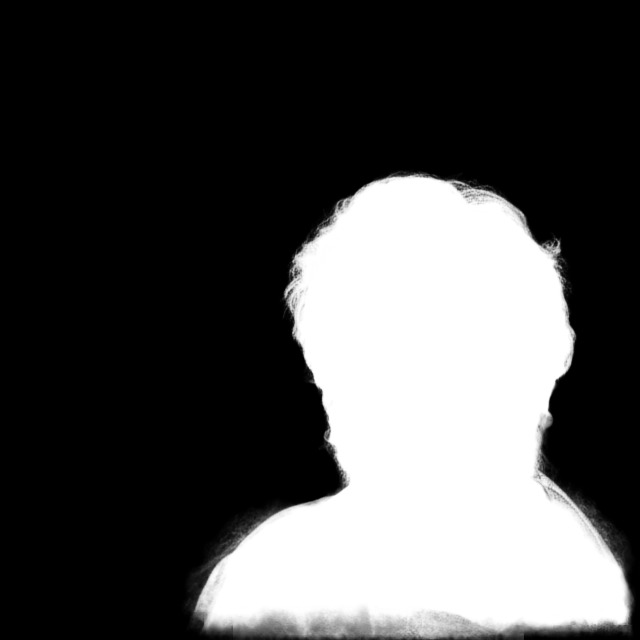}}{\footnotesize Alpha matte 1}\hspace{4pt}
    \stackunder[2pt]{\includegraphics[height=2.5cm,trim=15 15 0 0,clip]{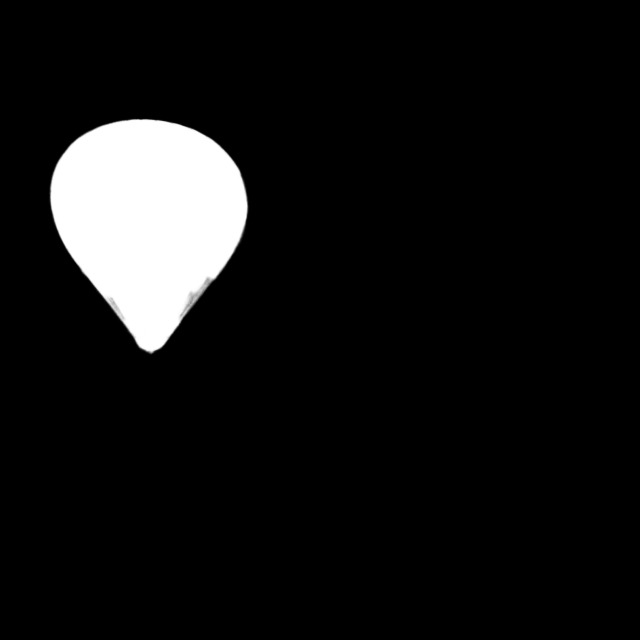}}{\footnotesize Alpha matte 2}\\[0.5ex]
    
    \stackunder[2pt]{\includegraphics[height=2.5cm,trim=15 15 0 0,clip]{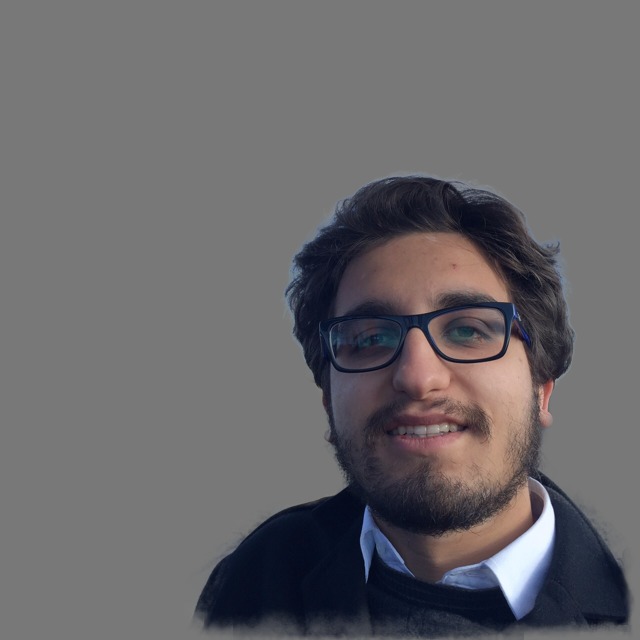}}{\footnotesize Instance 1}\hspace{4pt}
    \stackunder[2pt]{\includegraphics[height=2.5cm,trim=15 15 0 0,clip]{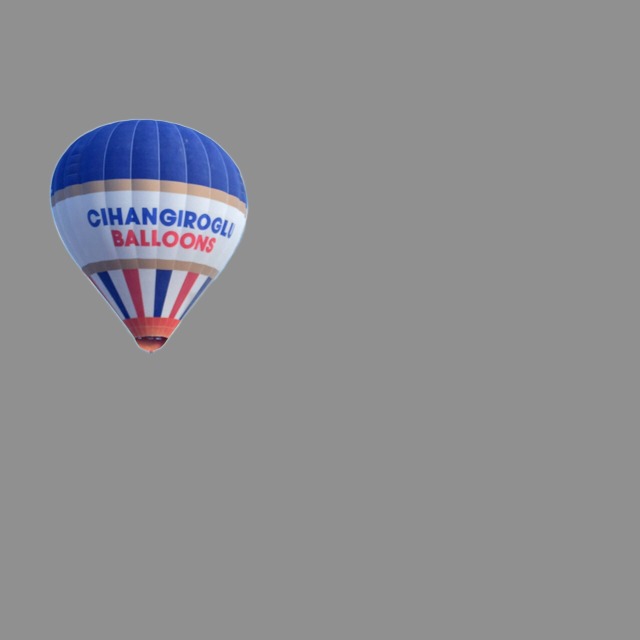}}{\footnotesize Instance 2}\hspace{4pt}
    \stackunder[2pt]{\includegraphics[height=2.5cm,trim=15 15 0 0,clip]{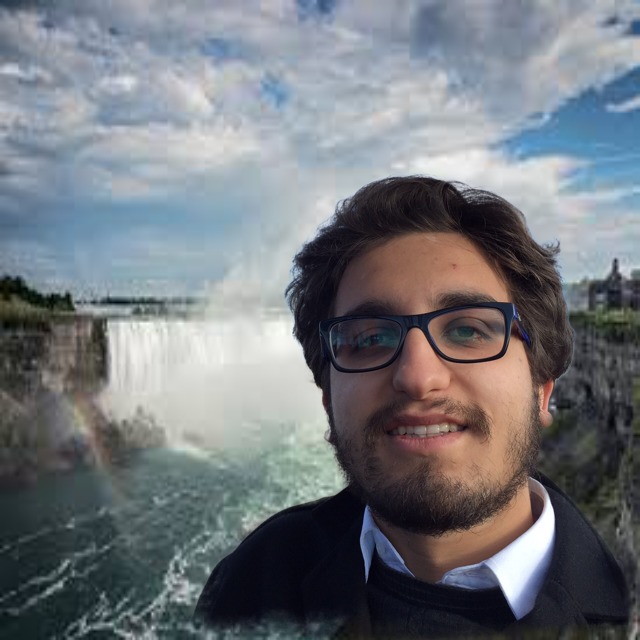}}{\footnotesize Composite 1}\hspace{4pt}
    \stackunder[2pt]{\includegraphics[height=2.5cm,trim=15 15 0 0,clip]{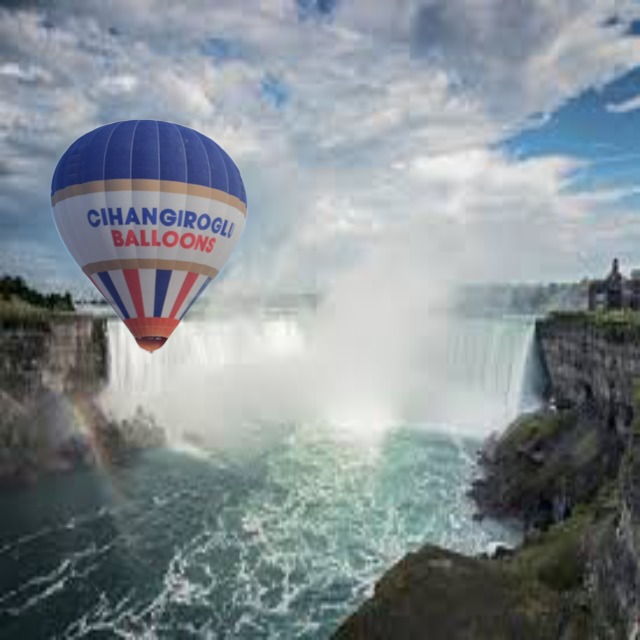}}{\footnotesize Composite 2}\hspace{4pt}
    \stackunder[2pt]{\includegraphics[height=2.5cm,trim=15 15 0 0,clip]{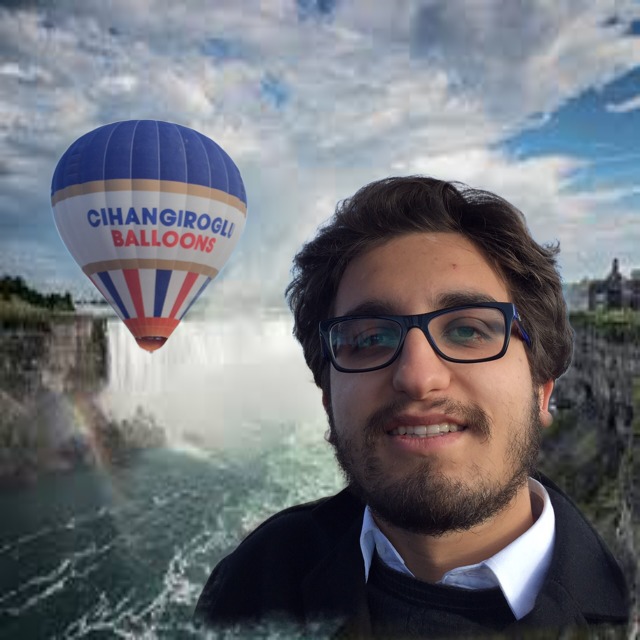}}{\footnotesize Composite all}\\[-1ex]
    \captionof{figure}{Our approach first generates coarse instance masks that are used to create trimaps. Then, the trimaps and the original image are used by the image matting network to produce an alpha matte for each instance. Finally, the alpha mattes are used for image compositing.}
    
    \label{fig:fig1}
\end{center}%
}]

\begin{abstract}
Image compositing is a key step in film making and image editing that aims to segment a foreground object and combine it with a new background. Automatic image compositing can be done easily in a studio using chroma-keying when the background is pure blue or green. However, image compositing in natural scenes with complex backgrounds remains a tedious task, requiring experienced artists to hand-segment. In order to achieve automatic compositing in natural scenes, we propose a fully automated method that integrates instance segmentation and image matting processes to generate high-quality semantic mattes that can be used for image editing task. 


Our approach can be seen both as a refinement of existing instance segmentation algorithms and as a fully automated semantic image matting method. It extends automatic image compositing techniques such as chroma-keying to scenes with complex natural backgrounds without the need for any kind of user interaction. The output of our approach can be considered as both refined instance segmentations and alpha mattes with semantic meanings. We provide experimental results which show improved performance results as compared to existing approaches.

\end{abstract}


%
\IEEEpeerreviewmaketitle

\section{Introduction}
Image compositing is a key technique in movie production and image editing. It combines visual elements from different sources into one image. The process of extracting visual elements from a source typically involves precise extraction of foreground objects from the background using either manual rotoscopy, where an artist traces the object to be extracted, or by an automatic chroma-keying when the object is in front of a uniformly colored background (e.g. green-screen). In this work, we present a method for automatic foreground object extraction that can work even with complex backgrounds. Our method combines instance segmentation and image matting processes, which allows for multiple foreground objects to be classified and segmented and extracted from the background. The semantic labels provided by the instance segmentation process provides a way to automatically extract objects of different types. For example, we can extract all people in the scene from the background, leaving other types of foreground objects behind. Or, we can obtain multiple mattes, one for each instance, and these can be used as desired in later compositing operations.

Object segmentation is considered as one of the most important and complex tasks in Computer Vision. There have been many recent advances in segmentation algorithms based on learning-based methods, in large part due to a number of publicly available datasets that provide human-drawn segmentation masks as ground truths. However, most of these masks are coarse and unrefined, which makes the segmentation algorithms also provide coarse and unrefined object boundaries. These image segmentation techniques lack sufficient refinement to be used for high quality image composition tasks.

On the other hand, image matting is another fundamental problem in Computer Vision that has been studied since the 1950s. Similar to segmentation, which involves generating a coarse binary mask for each object, image matting extracts an interesting object from a static image or a set of video frames by estimating an alpha matte $\alpha$ containing the opacity value for each pixel in the image. Typically, pixels in the foreground have alpha values equal to 1 while those in the background are given alpha values equaling 0. However, because of the extended size of pixels as well as motion blur in videos, pixels on the boundary between foreground and background objects have contributions from both the foreground and background objects and hence are given alpha values between 0 and 1. This relation is interpreted in Equation \ref{alpha}, where $i, FG, BG$ represent pixel position, foreground image, and background image respectively. 
\begin{equation}\label{alpha}
	I_i = \alpha_i\times FG_i + (1-\alpha_i)\times BG_i
\end{equation}

\begin{figure}[ht]
\centering
\captionsetup{justification=centering}
\subfloat[RGB input]{\includegraphics[height=1.25cm]{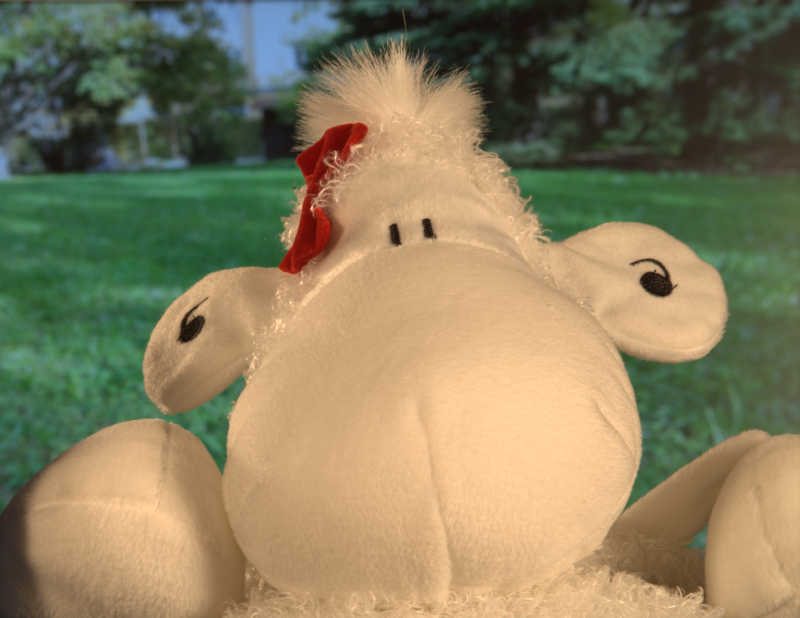}\label{f1a}}
\subfloat[Trimap]{\includegraphics[height=1.25cm]{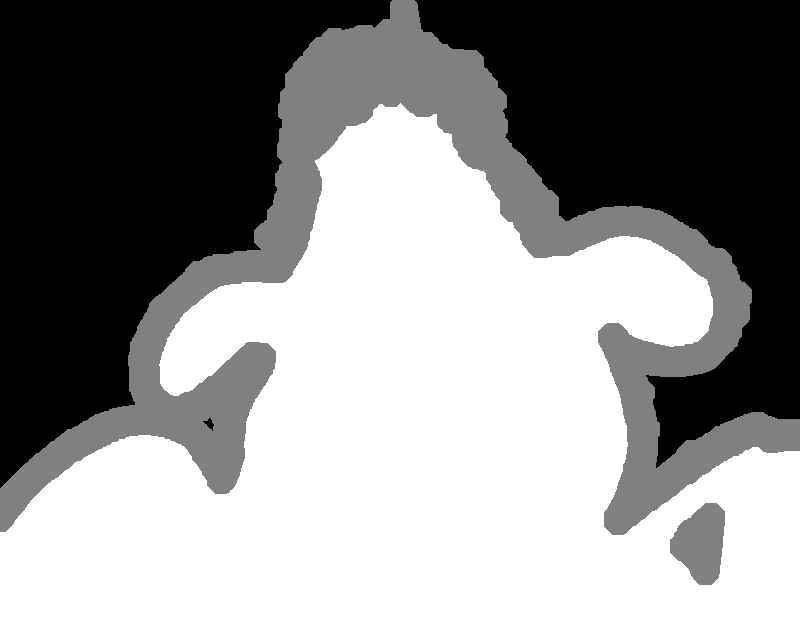}\label{f1b}}
\subfloat[Strokes]{\includegraphics[height=1.25cm]{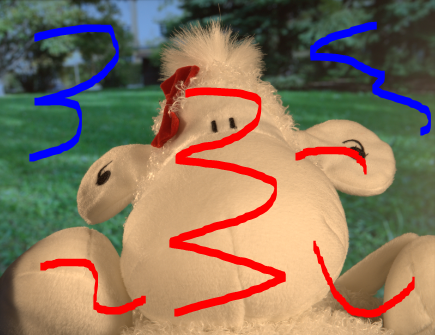}\label{f1c}}
\subfloat[Spectral features]{\includegraphics[height=1.25cm]{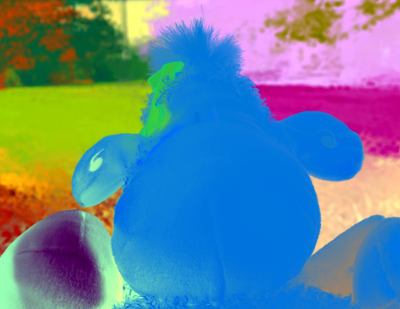}\label{f1e}}
\subfloat[Alpha matte]{\includegraphics[height=1.25cm]{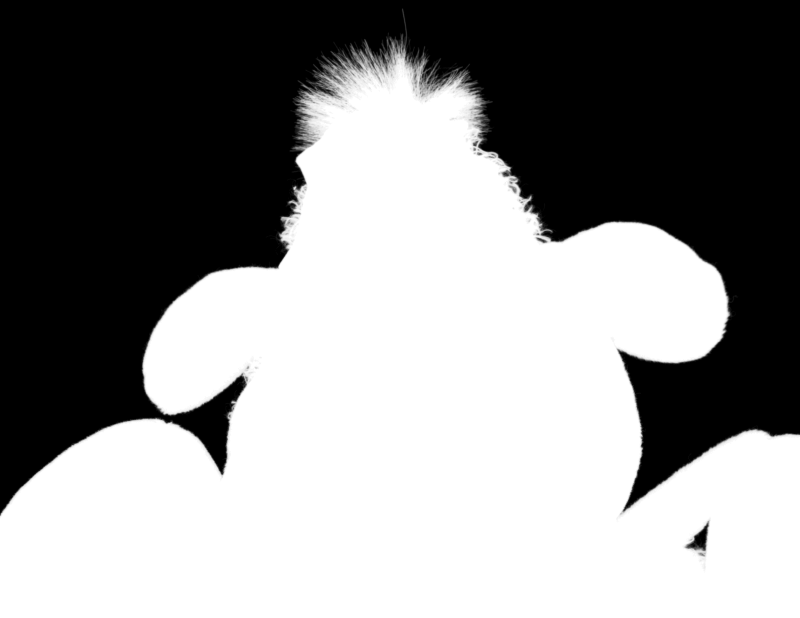}\label{f1d}}

\caption{Image matting input and output indication. Spectral features in (d) requires user guidance to select foreground features.}
\end{figure}

In equation \ref{alpha}, the only known value is the image input $I$, while the variables $FG, alpha, BG$ are unknown and need to be estimated. To simplify the estimation process, most image matting algorithms require a manual intervention in the form of user-labeled inputs. One type of user-labeled input is called the \emph{trimap} (Figure\ref{f1b}), which densely labels the opacity of the known (foreground and background) and unknown (boundary) regions. In the trimap, the foreground and the background are labeled $\alpha=1$ and $0$, respectively, while the remaining regions are initially given a label of $\alpha_i=0.5$. Another common type of user-labeled input are \emph{strokes} (Figure\ref{f1c}), which labels the region of foreground and background coarsely using scribbled strokes. Stroke-based algorithms are faster, requiring less user input, but produces lower quality results. Finally, there exist semi-automated methods for generating matting features directly from RGB input (e.g. Levin's spectral matting approach [11]) as shown in Figure \ref{f1e}. These techniques are considered semi-automated since they still require a small amount of user guidance to select the corresponding foreground features. 
The requirement of user interaction not only causes major delays and expense in the image editing workflow, but also severely limits the applications that image matting can be used on. 

In this paper, we propose a fully automated image segmentation-matting approach in which accurate segmentation can be achieved on general natural image datasets such as COCO \cite{coco}. The proposed method takes an RGB image without any additional user-labeled input and generates an alpha matte for all the detected object instances in the scene. The proposed deep learning-based approach consists of two stages. First, the input image is fed into a Mask R-CNN network \cite{mask rcnn} where the object bounding box and instance mask are generated. Then, using these two results, a trimap is estimated for each detected object in the image. Next, the trimap and the original RGB image are used together as inputs to a Deep Image Matting network \cite{deep image matting} to generate the final alpha matte. The final output of the pipeline is not only the refined foreground object mask but also an alpha matte with an accompanying semantic label that can be used to extract specific types of objects in a complex scene. 

We evaluate our technique qualitatively by comparing results with other approaches tackling automatic matting or automatic soft segmentation problems. Experimental results demonstrate comparable and, in some cases, superior, performance in extracting alpha mattes as compared with existing approaches. Our method is also capable of compositing with the motion blur that is often encountered in movies and videos so that the new background pixels show through on the blurred region. Our approach allows non-experts to perform natural image compositing easily by just selecting the object categories to keep and the auto-generated semantic alpha mattes will do the rest. It will also accelerate video compositing tasks that originally required artists to laboriously trace objects frame by frame. An example of using our approach in image compositing can be seen in Figure \ref{fig:fig1}, including the intermediate trimaps and instance segmentations. 

\begin{figure*}[ht]
        \captionsetup{justification=centering}
        \centering
        \includegraphics[scale = 0.65]{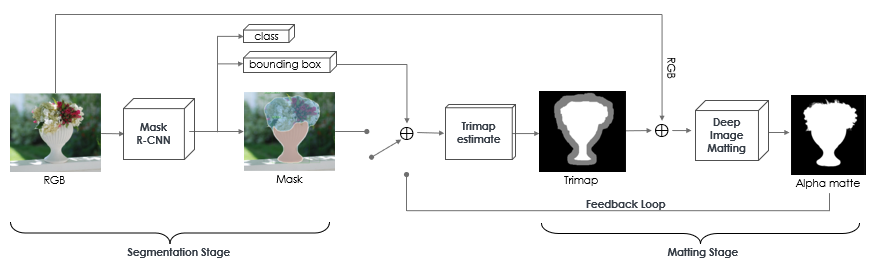}\\[-3ex]
        \caption{Proposed method as a pipeline}
        \label{arch}        
\end{figure*}

\section{Related Work}
\subsection{Natural Image Matting}
Image matting estimates an alpha matte for an interesting object by generating an opacity value for each pixel in the image. A trimap is a typical input to matting algorithms which indicates uncertain or unknown opacity (boundary) regions that the algorithms should work on, as well as known foreground and background regions. Matting methods can be categorized into: Non-learning-based and Learning-based methods. There are many non-learning image matting methods such as sampling-based \cite{sample-based-matting}, Bayesian-based \cite{sample-based-matting2}, affinity-based methods like Closed-form Matting \cite{closed-form}, and Poisson Matting\cite{shared-sampling}. Learning-based methods were only recently developed because of the difficulty in generating large scale ground truth data. Xu et al. used deep learning in solving the matting problem by creating a new dataset with real world images and expanded the dataset scale by compositing it with different background images. They use a two-stage deep learning method consisting of a deep convolutional encoder-decoder network and a fully convolutional network for refinement\cite{deep image matting}. Using the same dataset, \cite{alphagan} replaces the refinement stage in \cite{deep image matting} with adversarial model, achieving state-of-the-art performance on benchmark. Learning based methods have the advantage that the semantic meaning of the objects are learned, thus performing better in the tricky cases, for example, when the foreground and background colors are similar.

\subsection{Instance Segmentation} 
Instance Segmentation is a combination of object detection and semantic segmentation. Object detection classifies each object into a category and localize each object by predicting a bounding box. Semantic segmentation classifies each pixel in the image into an object category without distinguishing different instances. Mask R-CNN \cite{mask rcnn} is a state-of-the-art instance segmentation method. It takes an RGB image as input and generates three types of annotations for each detected object. The common dataset to train Mask R-CNN is COCO dataset \cite{coco} containing 81 object categories. For each recognized object in the input image, Mask R-CNN returns an object label, a bounding box indicating object location and scale, and a segmentation mask specifying the object in pixel level.

\subsection{Soft Segmentation}
Soft segmentation, in contrast to traditional segmentation that focuses on generating coarse object masks, estimates more precise object boundary transactions. Semantic Soft Segmentation \cite{soft_seg} is a state-of-the-art work achieving good performance on automatic soft segmentation. It uses high-level semantic features extracted from the semantic segmentation model DeepLab \cite{deeplab} to categorize and combine low-level texture and color features generated from spectral decomposition.  

\section{Methodology and Experimental set-up}\label{methods}
This section introduces the pipeline of the proposed method (Figure \ref{arch}) and how each key part works individually and together. The pipeline starts with taking an input image into a Mask R-CNN network through which the object bounding box and instance mask are generated. Using these intermediate outcomes, a coarse trimap with large uncertainty region can be estimated for each detected object. To estimate the final alpha matte, this trimap and the input image are fed to the image matting network. Due to the coarseness of the instance mask leading to low-quality trimaps, the alpha matte generated in the first pass is also poor. Estimating a new trimap from the generated alpha matte and passing through the matting stage again generally improves the segmentation and matting results. This forms a feedback loop in the pipeline where the trimap and alpha matte improve each other.

\subsection{Segmentation stage}

The first stage of the pipeline could use any instance segmentation network that produces instance masks and bounding boxes. We use the Mask R-CNN \cite{mask rcnn} algorithm in our approach. As the first stage of the pipeline, the inference error of Mask R-CNN influences the accuracy of the latter steps. Therefore, we use \emph{Mask R-CNN with bells and whistles} that achieves higher accuracy in both bounding box and mask inference. The model is built on a deeper backbone architecture ResNeXt-152-32x8d-FPN trained on ImageNet-5k as opposed to the usual ImageNet-1k \cite{github-maskrcnn}. We use the pre-trained Mask R-CNN model from Detectron Pytorch version \cite{github-maskrcnn}.


\subsection{Trimap estimation stage}
Trimap estimation makes use of the intermediate outputs from Mask R-CNN to produce a trimap for each detected object instance. We make an assumption that the region near the mask boundary is the region that requires the most further estimation, i.e. the unknown region where the image matting algorithm should focus on. Therefore, a certain region dilated from the object mask is defined to be the unknown area in the trimap with $\alpha_i=0.5$. The region further inside the mask defines the foreground area with $\alpha_i=1$, while the region further outside the mask is assigned as the background with $\alpha_i=0$.

The amount of the dilation is determined by the object size. The approximate width and height of an object can be estimated using its bounding box coordinates as $width=bbox[2] - bbox[0]$ and $height=bbox[3] - bbox[1]$. We choose the dilation rate to be a fixed percentage of the width-and-height average. There is a trade-off in choosing a higher or lower dilation rate. A precise trimap is favored by the matting network as it imposes stronger constraints. When the mask boundary is close to the object true boundary, a small rate of dilation is enough to cover the region that needs to be refined. But there are also cases when larger dilation rate is preferable to recover errors of poor object mask, resulting in larger uncertainty region in the trimap that could degrade the matting stage.  




\subsection{Matting stage}
For image matting,  we use the learning-based method \emph{Deep Image Matting} by Lin et al.\cite{deep image matting}. It is a VGG16-based encoder-decoder network followed by a fully-connected refinement stage. Taking advantage of the semantic meaning extracted by VGG16, Deep image matting has an outstanding performance on natural scenes where the foreground color and background color are sometimes very similar. Its performance is also less dependent on the quality of the trimap than non-learning methods. This data-driven model uses the dataset Lin et al. created \cite{deep image matting}. The dataset contains 431 unique objects with associated alpha mattes but the object variety is still limited compared to common object datasets (with no mattes) used for training object classifiers. 

\subsubsection{Network training}
We only train the network without the last refinement layers due to hardware limitations. We expect our results to be even better with the refinement layers. The implementation uses the dataset from Lin et al. \cite{deep image matting}.
Training data of the network is pre-processed as $320\times320$ image patches centering at an unknown pixel (i.e. $\alpha_i=0.5$) randomly cropped from training images.

\subsubsection{Inference}
During test time, we pre-process the test images similarly as in training in order to achieve better performance. When dealing high resolution test images we could simply resize them (in our case to $320\times320$). However, this may cause unavoidable issues in terms of network performance and coarse alpha mattes. Indeed, downsampling methods causes detail loss especially when the ratio of the size of test images to the desired size is very high.
To avoid this, we implemented a \emph{patch-based pre-processing method} on the input test images. This approach consists in cropping a test image into patches of $320\times320$. Each patch of that single image is separately fed to the network giving multiple alpha matte results. Then, these results are blended together. A limitation in using this technique is the fact that we must make sure that the cropped input images are centered at an unknown pixel in the trimap. If this is not the case the network may perform poorly. In addition to that, the network performance is compromised when the content of the cropped patch is very different from the training data used during the training phase of the network.
To address these issues, the high resolution test image is first downsampled to $640\times640$. Then, patches are randomly cropped with their centers moving along the gray region of the trimap until the whole unknown region gets covered. For each patch, an alpha matte patch is generated. All these alpha matte patches are pasted back to their original locations and averaged with the existing overlaps of patches. With this method of averaging we found some issues related to discontinuities especially along the borders as well as susceptibility to outliers. Hence, we developed a \emph{multiple Sampling method} which consists in running the test algorithm on each image for $K=10$ times and take the median of each set of alpha values at a pixel to exclude any possible outliers and proceed in smoothing the final alpha matte.

\subsection{Feedback loop}
Since the trimap generated directly from the instance mask is generally coarse, the alpha matte from the first pass of the pipeline is normally undesirable with low boundary accuracy. So, a feedback connection was inserted between the alpha matte output and the image matting network input. From this, a new trimap is created by dilation from the previous pass's alpha values. The latter is fed back to the image matting network for further refinement. The motivation behind this implementation is that the quality of trimap and alpha matte increases at each pass and leads to a simultaneous improvement. With this increase in quality, the dilation rate of the trimap is decreased. Generally, four feedback loops leads to accurate results as demonstrated in Section \ref{Evaluation}. 

\subsection{Handling multiple objects }
When faced with an input image that contains multiple objects, we proceed in dealing with one object at a time (i.e. in a case of two objects in one image we consider the first object as foreground and the second as background image, then vice-versa). This method copes with superposing regions of the trimap after performing dilation of the boundaries.  Moreover, since the matting network used in this work is one designed for single-object matting, this method of distinction of object within a single image is more effective.

\section{Experimental Analysis}\label{Evaluation}
Natural image matting results can be evaluated numerically if the ground truth alpha matte is available. 
However, the purpose of our method is to generalize natural image matting to common datasets (e.g COCO) so that matting techniques can adapt to more general applications. In these datasets, ground truth alpha matte is not available. Therefore, this section mainly focuses on analysing qualitative results. We first analyse the effect of the pipeline structure that successfully refines coarse instance mask to alpha matte that can be used for image editing task. Then we compare results with existing fully automate methods including Semantic Soft Segmentation and Spectral Matting. In the end, we present some limitations and failure cases. 

\begin{figure}[ht]
\centering
\captionsetup{justification=centering}
\subfloat[RGB input]{\includegraphics[height=2.4cm,trim=100 35 100 15,clip=true]{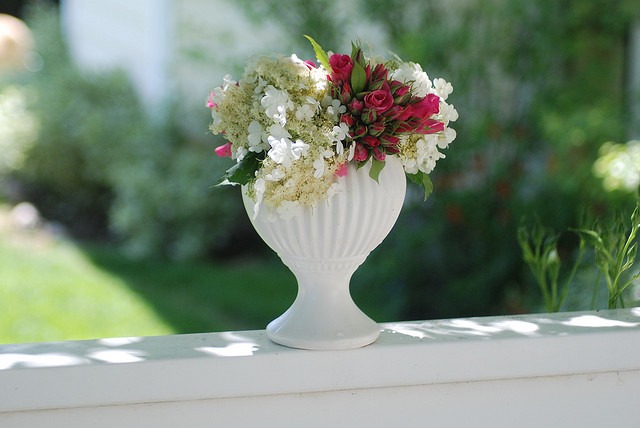}\hspace{3pt}}
\subfloat[Mask RCNN]{\includegraphics[height=2.4cm,trim=40 15 40 10,clip=true]{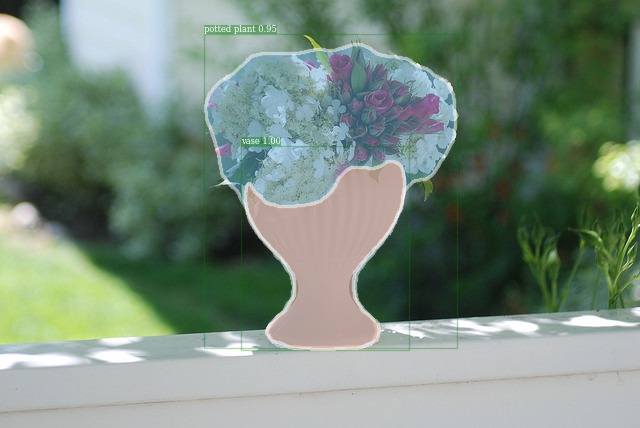}\hspace{3pt}}
\subfloat[Trimap, no feedback]{\includegraphics[height=2.4cm,trim=100 35 100 15,clip=true]{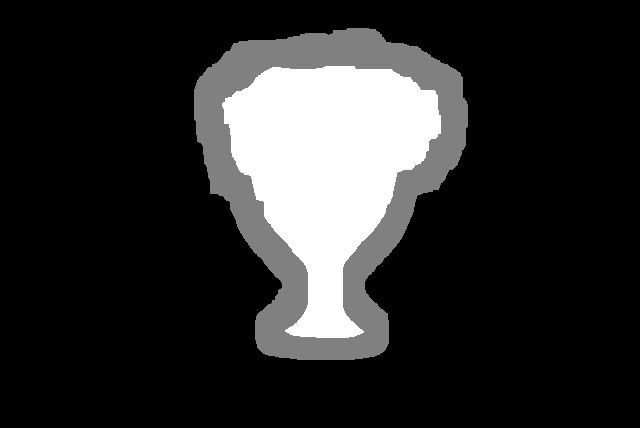}\label{trimap0}}\\[-1.5ex]
\subfloat[Alpha, no feedback]{\includegraphics[height=2.4cm,trim=100 35 100 15,clip=true]{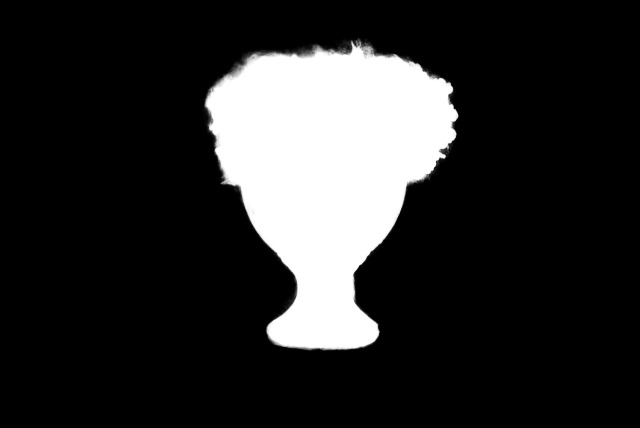}\label{alpha0}\hspace{3pt}}
\subfloat[Trimap after $2^{nd}$ pass]{\includegraphics[height=2.4cm,trim=100 35 100 15,clip=true]{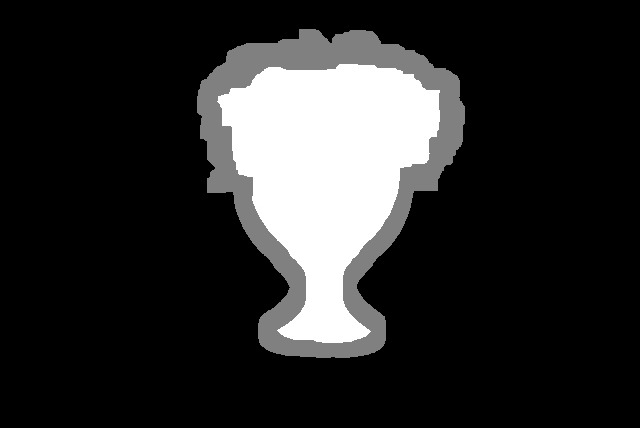}\label{trimap2}\hspace{3pt}}
\subfloat[Alpha after $2^{nd}$ pass]{\includegraphics[height=2.4cm,trim=100 35 100 15,clip=true]{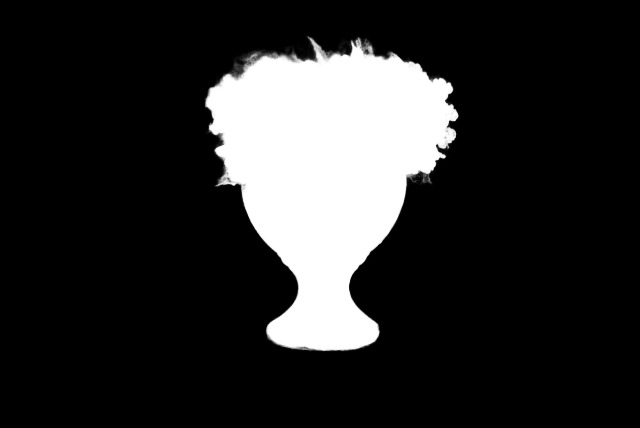}\label{alpha2}}\\[-1.5ex]
\subfloat[Trimap after $4^{th}$ pass]{\includegraphics[height=2.4cm,trim=100 35 100 15,clip=true]{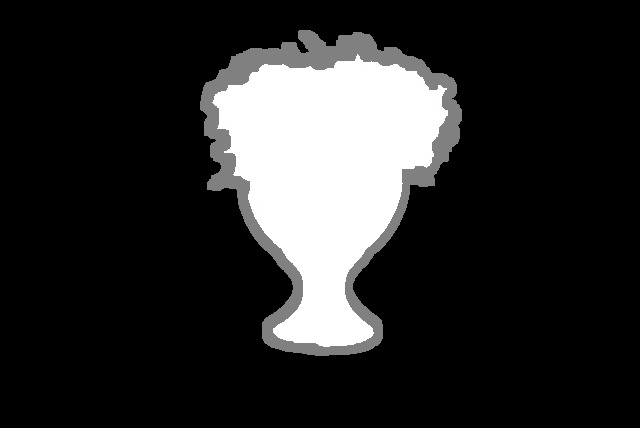}\label{trimap4}\hspace{3pt}}
\subfloat[Alpha after $4^{th}$ pass]{\includegraphics[height=2.4cm,trim=100 35 100 15,clip=true]{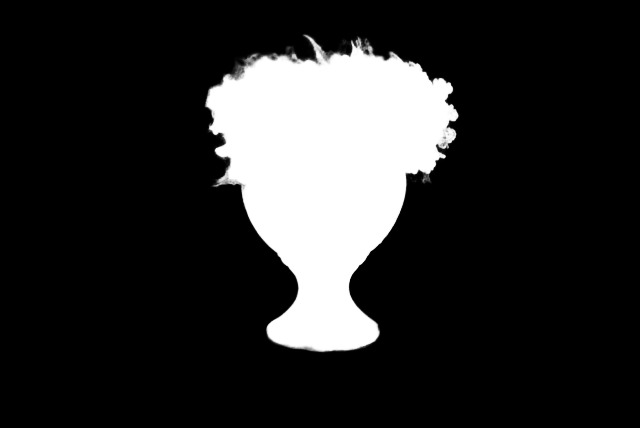}\label{alpha4}\hspace{3pt}}
\subfloat[Compositing result]{\includegraphics[height=2.4cm,trim=100 35 100 15,clip=true]{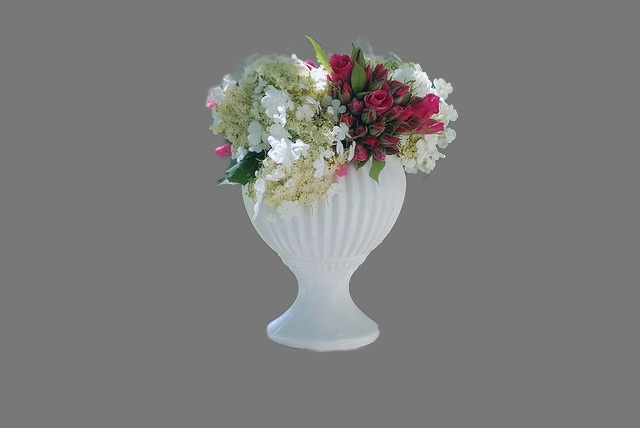}}\\[-1.5ex]
\caption{Matting results of potted plant image}
\label{plant}
\end{figure}
Figure \ref{plant} presents result details of an example image from the COCO dataset \cite{coco}. Mask RCNN detects the pot and the potted plant as two separate objects. Since the boundary structure of the pot is fairly simple, we only show the matting results of the potted plant. Figure \ref{trimap0} is the initial trimap generated directly from the instance segmentation mask with the widest and most coarse uncertain region. The alpha matte based on such a trimap, as shown in Figure \ref{alpha0}, lacks detail, but is slightly better than the original instance mask. A second trimap can be created by dilating this alpha matte, resulting in better trimap which then leads to better alpha matte. Figures \ref{trimap2}-\ref{alpha4} demonstrate the gradual improvement of trimaps and alpha mattes, where the number indicates at which feedback loop the corresponding result is generated from. A composite is created with the extracted plant region using the alpha matte generated after four feedback passes (Figure \ref{alpha4}). On a Nvidia Tesla K40 GPU, Mask RCNN takes an average of 39 seconds to generate separated trimap for each detected object and the matting stage takes 10 seconds with four feedback passes on each object.   

The number of feedback passes is a hyper-parameter that need to be adjusted based on the application requirements. We found that four feedback passes are generally enough to achieve high-quality performance. 

\subsection{Comparison}
As a comparison, segmentation results of the bouquet are also generated using Semantic Soft Segmentation \cite{soft_seg} and Spectral matting \cite{spectral} as shown in Figure \ref{plantcompare}. Both approaches suffer at the regions where the foreground object color is similar to the background as outlined in red. Both algorithms generate low-level features first as shown in the middle image and then gather the low-level features together using high-level features. Semantic Soft Segmentation takes advantage of high-level semantic meanings and groups low-level features into a few layers. However, there are still important details missing when the color of the foreground is too close to the background (e.g. around the most left leaves of the bouquet). 
\begin{figure}[ht]
\centering
\subfloat[Semantic Soft Segmentation results \cite{soft_seg}]{\includegraphics[height=2cm,width=8.5cm]{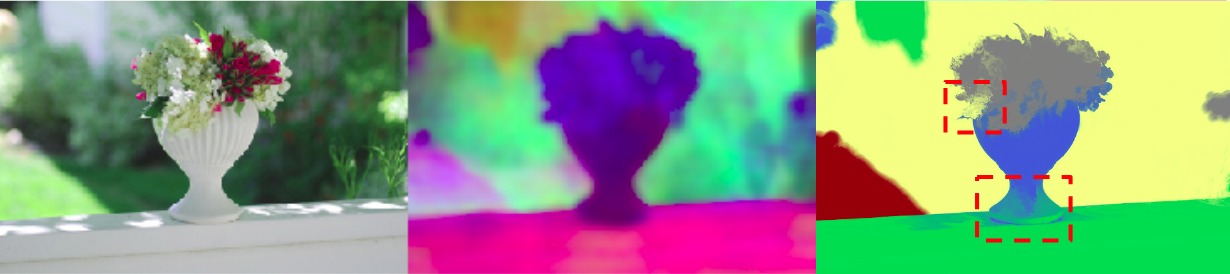}}\\[0.5ex]
\subfloat[Spectral matting results \cite{spectral}]{\includegraphics[height=2cm,width=8.5cm]{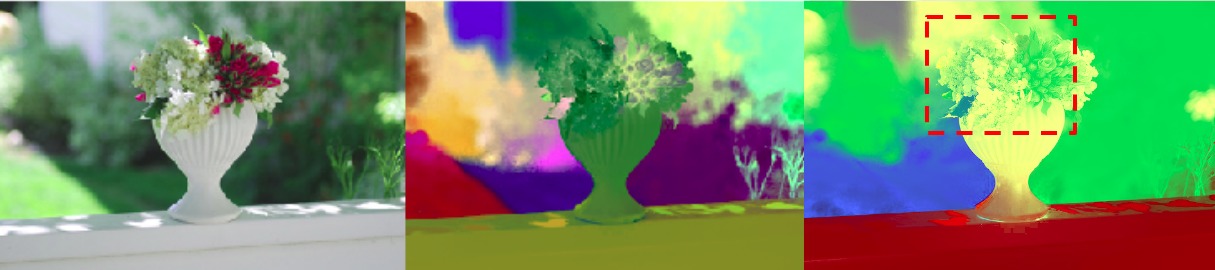}}\\[-1ex]
\caption{Results of Semantic Soft Segmentation and Spectral matting. From left to right: original image, extracted features, final result.}
\label{plantcompare}
\end{figure}

Results of more images from COCO dataset are presented in Figure \ref{resultsfig} and \ref{resultsfig2} including comparisons with Semantic Soft Segmentation and Spectral matting. The index on the left of each line in Figure \ref{resultsfig} and \ref{resultsfig2} will be used to refer to each presented image. Image 1-5 are the ones shown in the Semantic Soft Segmentation paper \cite{soft_seg}. The resolution of these images are slightly different than the originals in COCO. However, we expect the method to be robust to small image scale changes. For image 1-4 and 8, our approach achieves as good performance as Semantic Soft Segmentation. In image 4, our approach is the only one that correctly labels the black region on the top of the person's left shoulder as background. Although in image 8, the person's left hand is not detected, the composited result is still visually correct.   

Image 5 demonstrates our approach's capability of capturing motion blur like the person's right arm, while the other two methods cannot. Image 6, 7, and 9 demonstrate the advantage of using instance-level segmentation as guideline because when the scene contains more than one object that belongs to the same semantic category, only using semantic segmentation features tend to merge multiple different instances into one. Our approach is more flexible in choosing which instance to keep and this selection can also be automated by using the class label and bounding box position as instance identity. 

\begin{figure}
    \setcounter{figure}{5}
    \centering
    \captionsetup{justification=centering}
    \subfloat[RGB input]{\includegraphics[height=2.1cm]{cocoresults/COCO_train2014_000000001496_firehydrant_85_31.jpg}\hspace{3pt}}
    \subfloat[Our result]{\includegraphics[height=2.1cm]{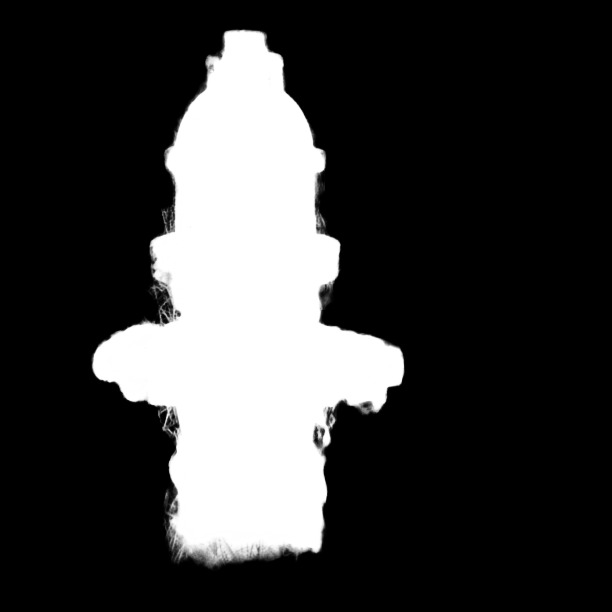}\label{hydrant-c}\hspace{3pt}}
    \subfloat[Our result, zoomed in]{\frame{ \includegraphics[height=2.05cm,trim=100 250 350 200,clip]{cocoresults/firehydrant-alpha-4.jpg}}\hspace{3pt}}
    \subfloat[Semantic Soft Segmentation]{\includegraphics[height=2.1cm]{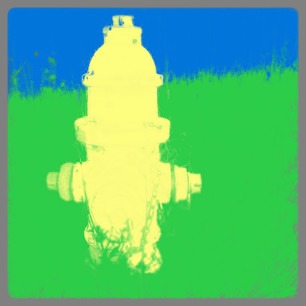}}\\[-1.5ex]
    \caption{Example of a failure case.}
    \label{hydrant}
\end{figure} 

\subsection{Limitations} 
One limitation of our approach is that the trimap generation is limited at object borders, hence for objects with large area of transparency or other type of complex opacity, the proposed method cannot generate trimaps covering such a large region, which results in incorrect alpha matte. The image matting network also suffers from the limited amount of object categories it is trained on. When applied to unfamiliar objects, the network tends to generate fuzzy and transparent alpha even though the truth object boundary is solid. As can be seen in Figure \ref{hydrant-c}, our approach confuses background grass as part of the fuzzy boundaries of the fire hydrant. The Semantic Soft Segmentation algorithm does not have the same issue, but has problems at the bottom of the hydrant.

\begin{figure*}[ht]
\setcounter{figure}{6}
\centering
\captionsetup{position=top}
\captionsetup{justification=centering}
\captionsetup[subfloat]{labelformat=empty}

\subfloat[RGB input]{\textbf{1}\hspace{3pt}\includegraphics[height=3cm]{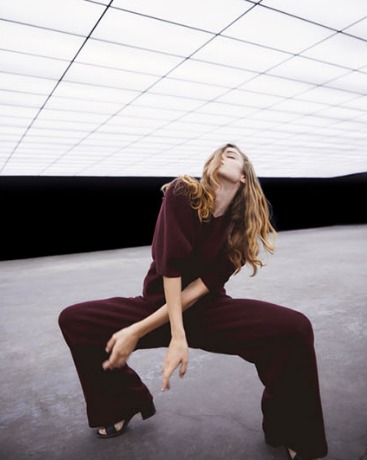}}
\subfloat[Mask R-CNN]{\includegraphics[height=3cm]{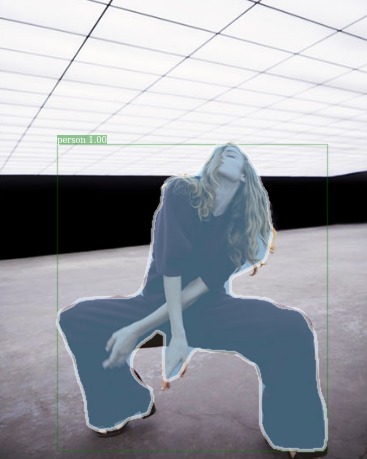}}
\subfloat[Trimap]{\includegraphics[height=3cm]{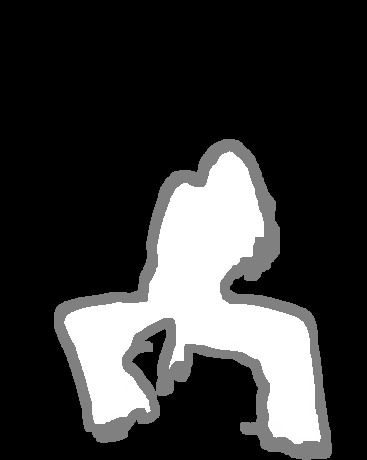}}
\subfloat[Alpha matte]{\includegraphics[height=3cm]{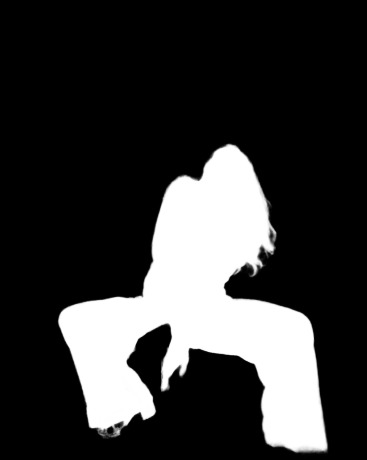}}
\subfloat[Instance compositing]{\includegraphics[height=3cm]{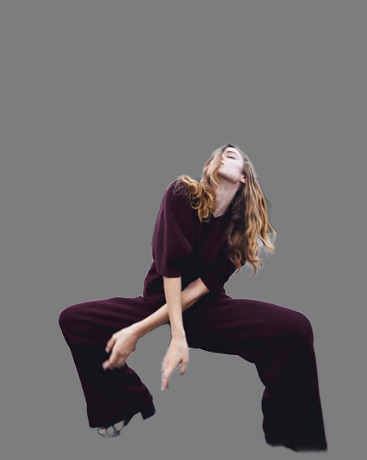}}
\subfloat[Semantic Soft Segmentation]{\includegraphics[height=3cm]{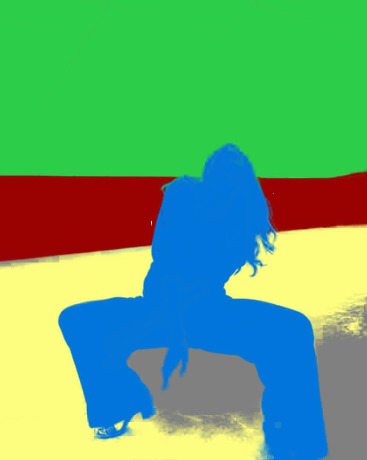}}
\subfloat[Spectral Matting]{\includegraphics[height=3cm]{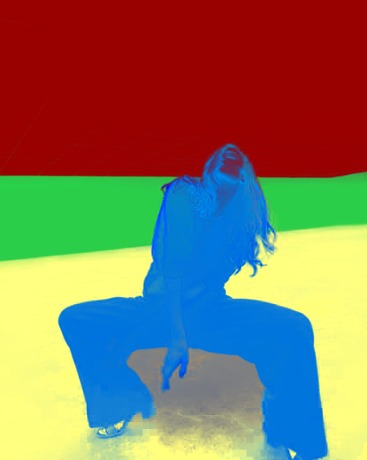}}

\subfloat{\textbf{2}\hspace{3pt}\includegraphics[height=3cm]{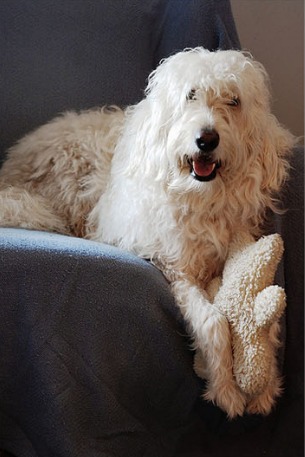}}
\subfloat{\includegraphics[height=3cm]{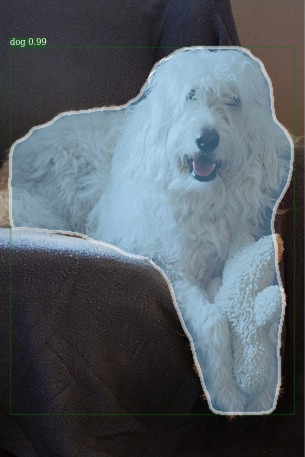}}
\subfloat{\includegraphics[height=3cm]{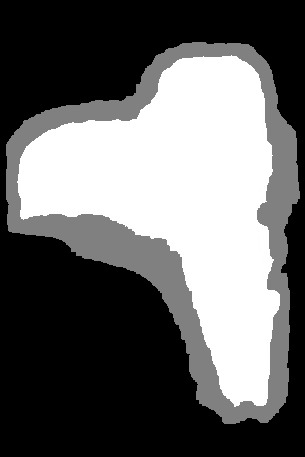}}
\subfloat{\includegraphics[height=3cm]{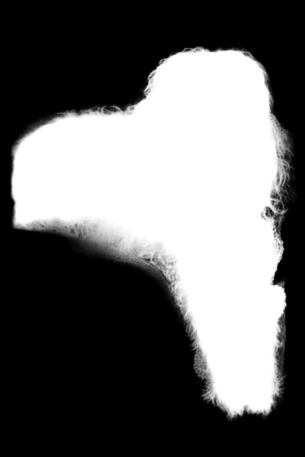}}
\subfloat{\includegraphics[height=3cm]{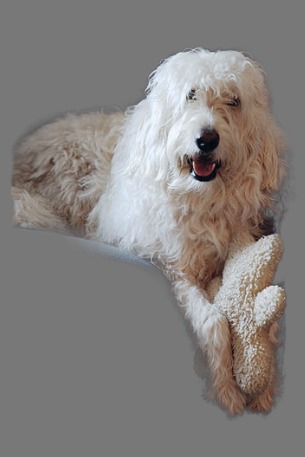}}
\subfloat{\includegraphics[height=3cm,trim=245 75 35 63,clip]{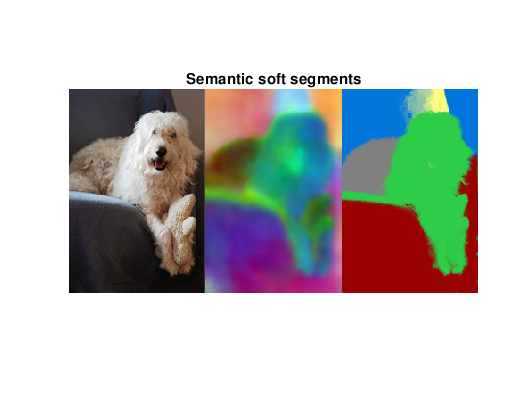}}
\subfloat{\includegraphics[height=3cm,trim=245 75 35 63,clip]{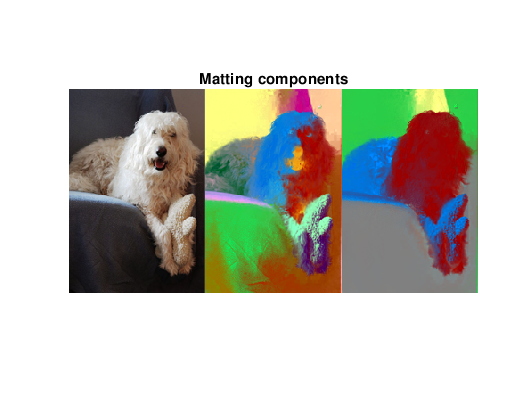}\hspace{3pt}}

\subfloat{\textbf{3}\hspace{3pt}\includegraphics[height=3cm]{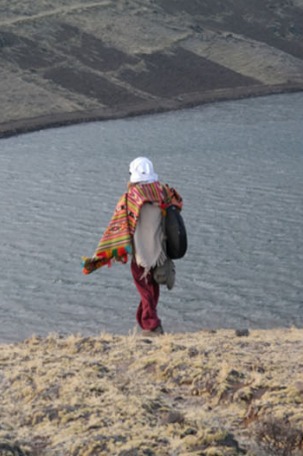}}
\subfloat{\includegraphics[height=3cm]{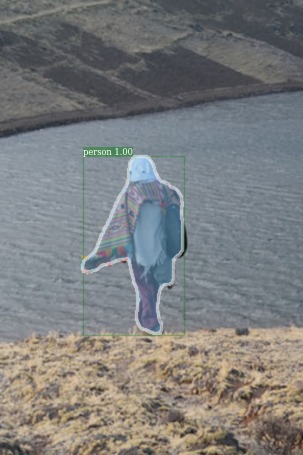}}
\subfloat{\includegraphics[height=3cm,trim=2cm 1.5cm 2cm 3cm,clip=true]{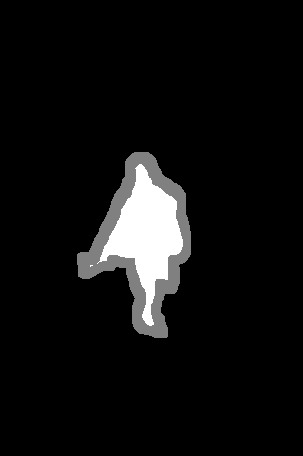}}
\subfloat{\includegraphics[height=3cm,trim=2cm 1.5cm 2cm 3cm,clip=true]{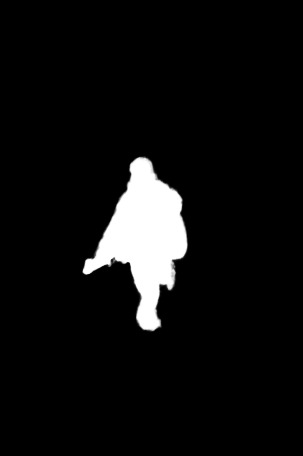}}
\subfloat{\includegraphics[height=3cm,trim=2cm 1.5cm 2cm 3cm,clip=true]{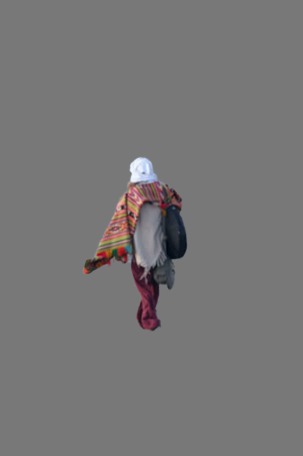}}
\subfloat{\includegraphics[height=3cm]{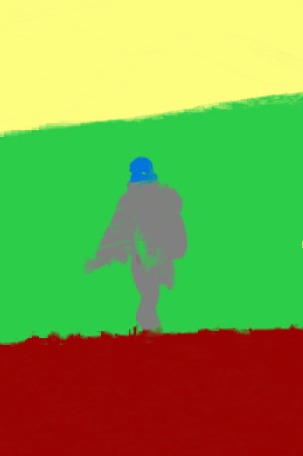}}
\subfloat{\includegraphics[height=3cm]{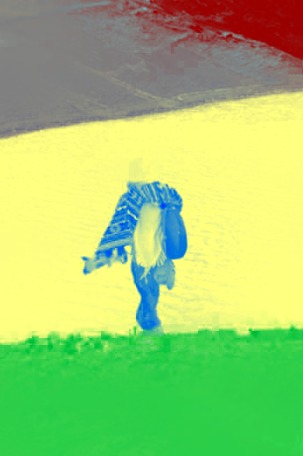}}

\subfloat{\textbf{4}\hspace{3pt}\includegraphics[height=2.5cm]{results/docia.jpg}}
\subfloat{\includegraphics[height=2.5cm]{results/docia-mask.jpg}}
\subfloat{\includegraphics[height=2.5cm]{results/docia_person_178_173trimap_4.jpg}}
\subfloat{\includegraphics[height=2.5cm]{results/docia_person_178_173alpha_4.jpg}}
\subfloat{\includegraphics[height=2.5cm]{cocoresults/composedocia_person_178_173alpha_4.jpg}}
\subfloat{\includegraphics[height=2.5cm]{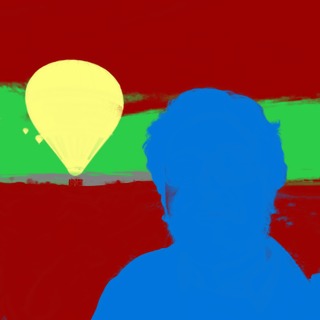}}
\subfloat{\includegraphics[height=2.5cm]{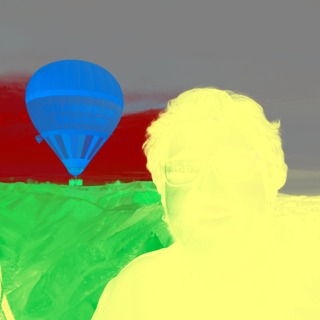}}

\subfloat{\textbf{5}\hspace{3pt}\includegraphics[height=3cm,trim=0 10 70 0,clip]{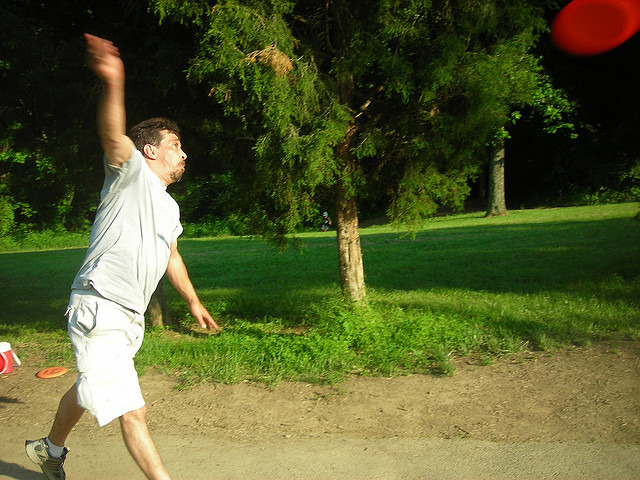}}
\subfloat{\includegraphics[height=3cm,trim=0 15 100 0,clip]{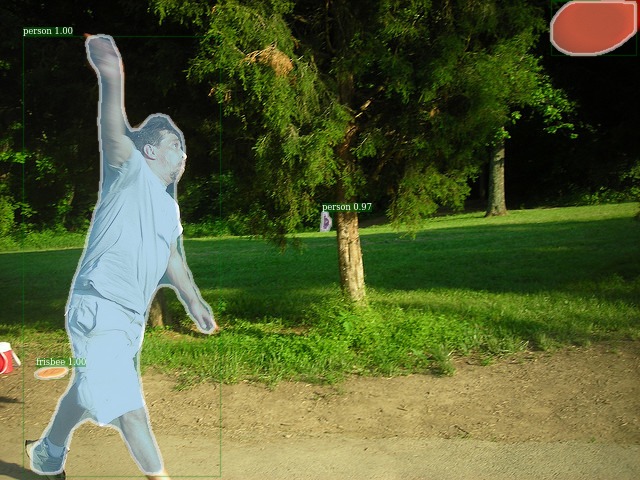}}
\subfloat{\includegraphics[height=3cm,trim=0 45 270 0,clip]{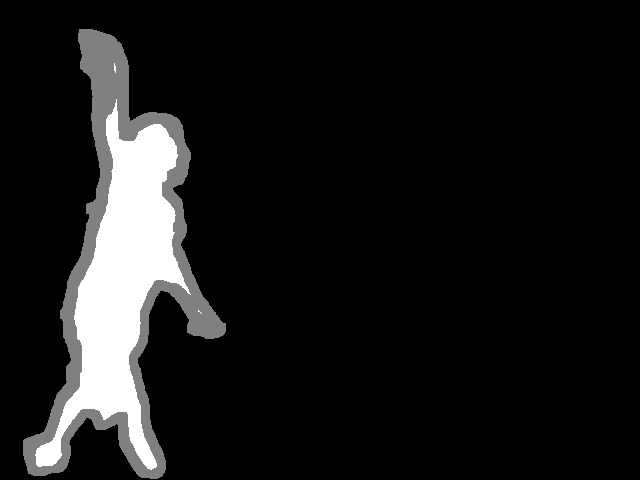}}
\subfloat{\includegraphics[height=3cm,trim=0 45 270 0,clip]{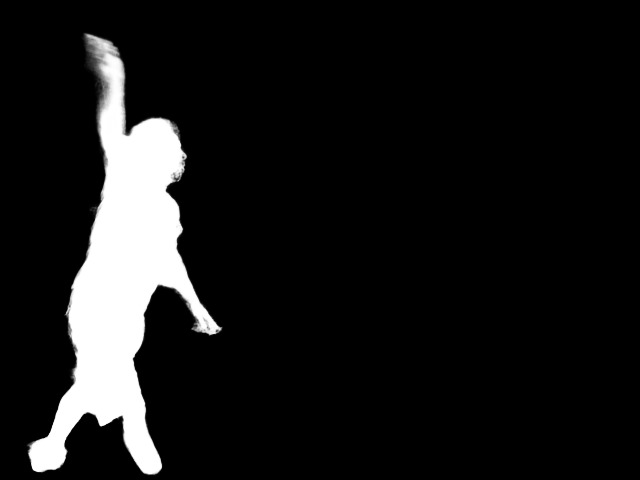}}
\subfloat{\includegraphics[height=3cm,trim=0 45 270 0,clip]{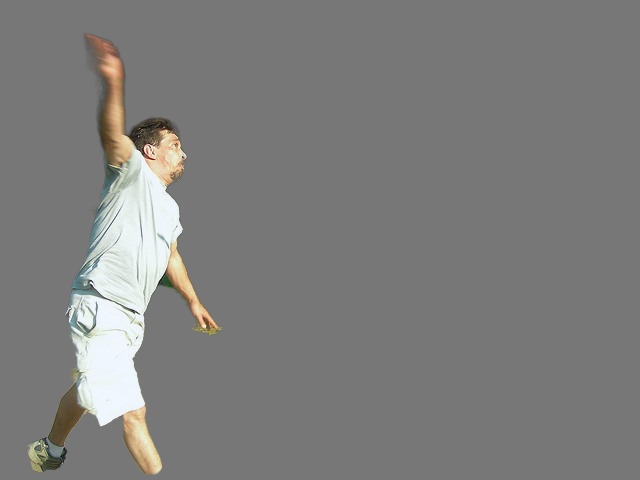}}
\subfloat{\includegraphics[height=3cm,trim=0 20 150 0,clip]{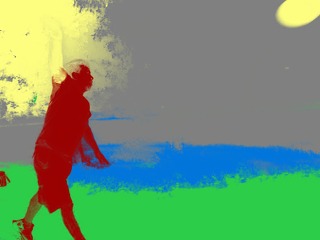}}
\subfloat{\includegraphics[height=3cm,trim=0 20 150 0,clip]{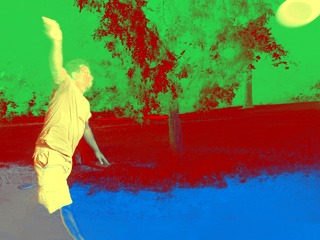}}

\subfloat{\textbf{6}\hspace{3pt}\includegraphics[height=2.5cm]{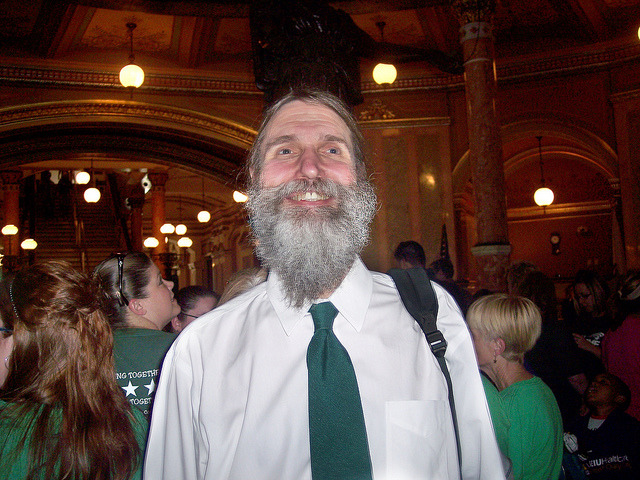}}
\subfloat{\includegraphics[height=2.5cm]{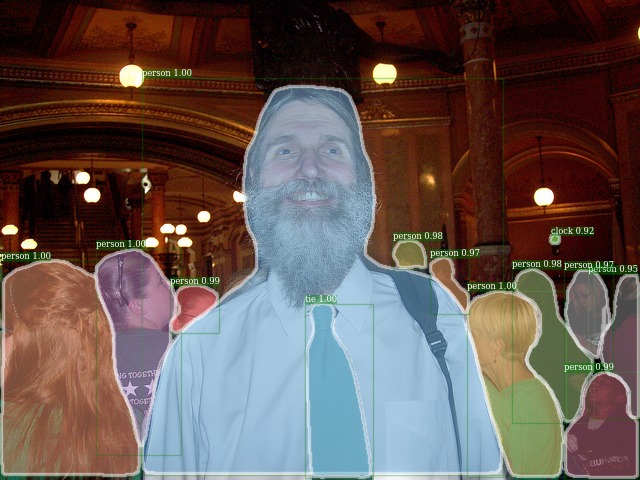}}
\subfloat{\includegraphics[height=2.5cm]{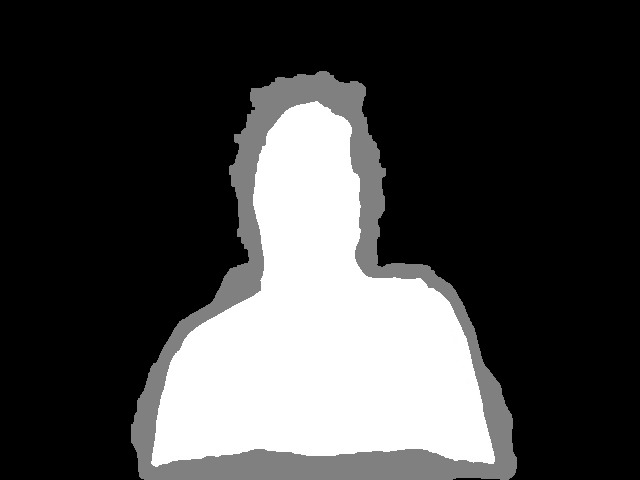}}
\subfloat{\includegraphics[height=2.5cm]{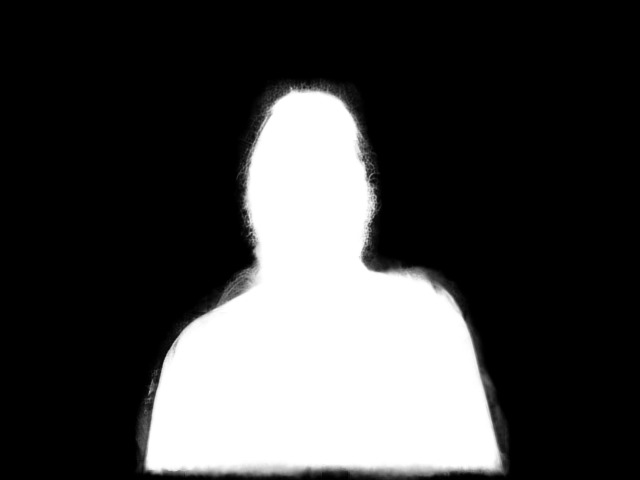}}
\subfloat{\includegraphics[height=2.5cm]{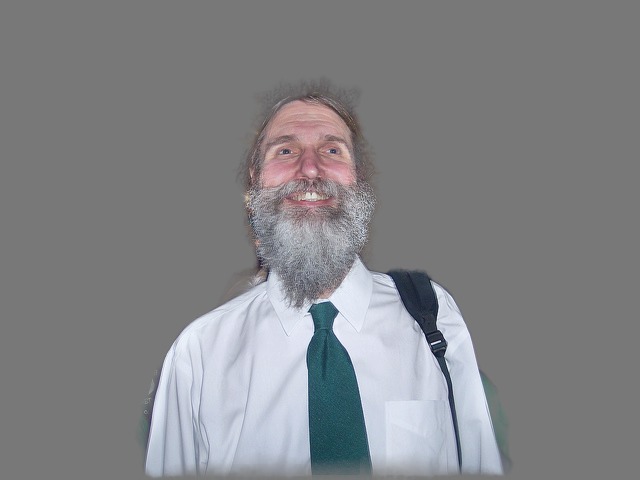}}

\subfloat{\includegraphics[height=2.5cm]{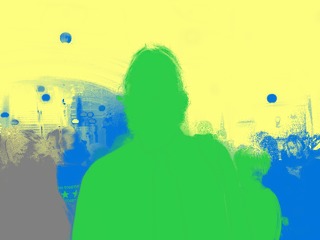}}
\subfloat{\includegraphics[height=2.5cm]{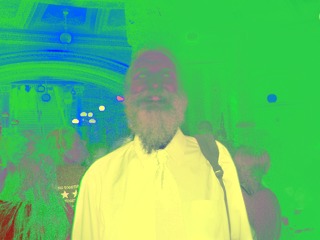}}

\caption{From left to right are RGB input image, Mask-RCNN, trimap at the 4th feedback loop, alpha matte at the 4th feedback loop, instance compositing, result of Semantic Soft Segmentation, and result of spectral matting.}
\label{resultsfig}
\end{figure*}

\begin{figure*}[ht]
\setcounter{figure}{7}
\centering
\captionsetup{position=top}
\captionsetup{justification=centering}
\captionsetup[subfloat]{labelformat=empty}

\subfloat[RGB input]{\textbf{7}\hspace{3pt}\includegraphics[height=2.7cm]{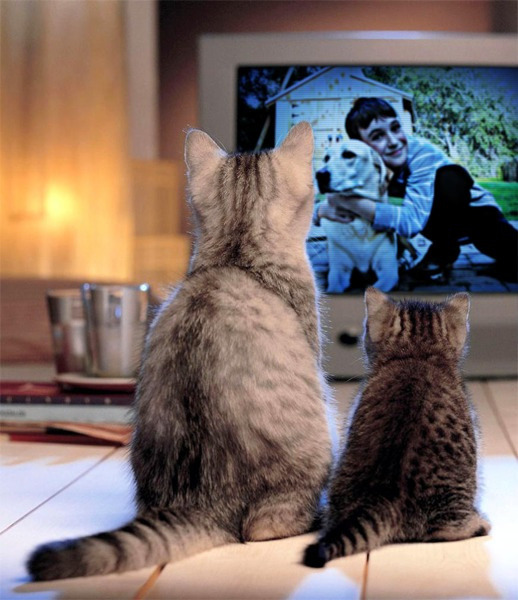}}
\subfloat[Mask R-CNN]{\includegraphics[height=2.7cm]{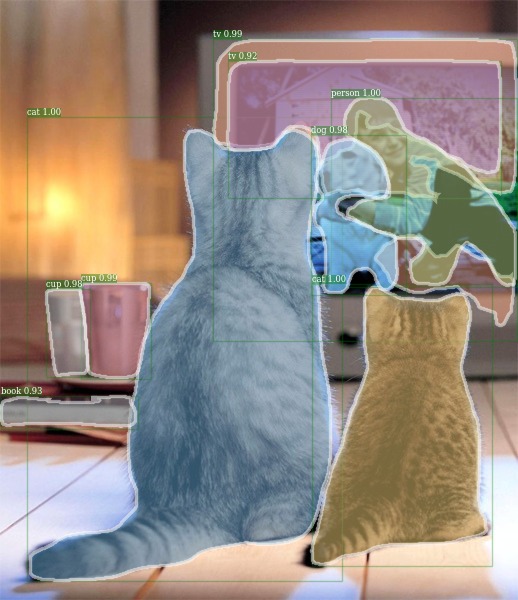}}
\subfloat[Trimap 1]{\includegraphics[height=2.7cm]{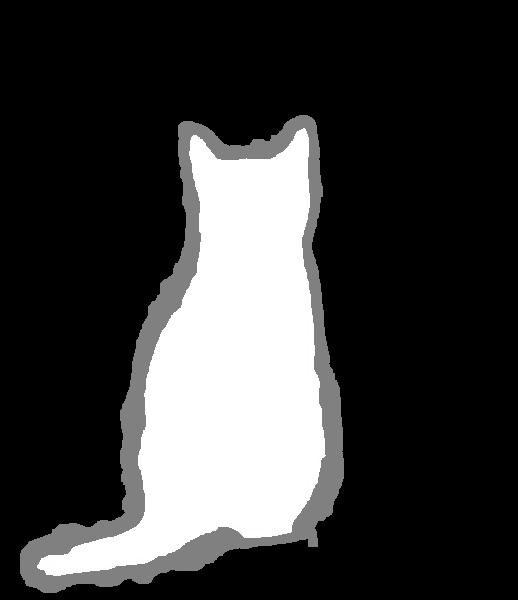}}
\subfloat[Trimap 2]{\includegraphics[height=2.7cm]{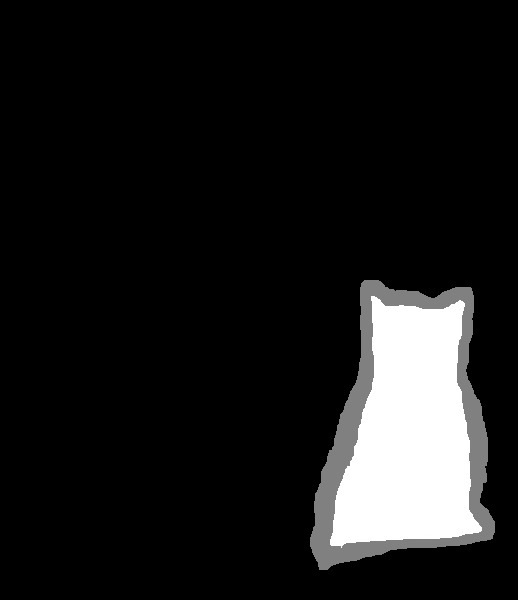}}
\subfloat[Alpha matte 1]{\includegraphics[height=2.7cm]{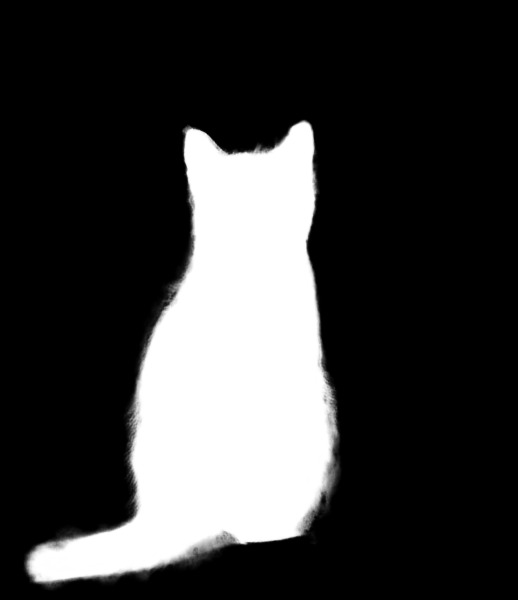}}
\subfloat[Alpha matte 2]{\includegraphics[height=2.7cm]{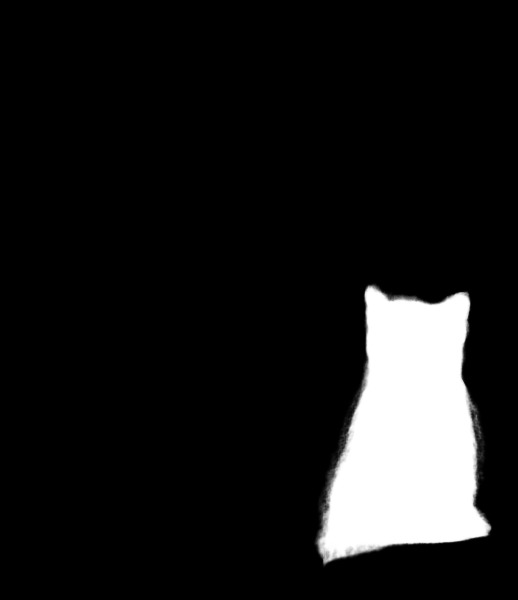}}\\
\subfloat[Instance compositing]{\includegraphics[height=2.7cm]{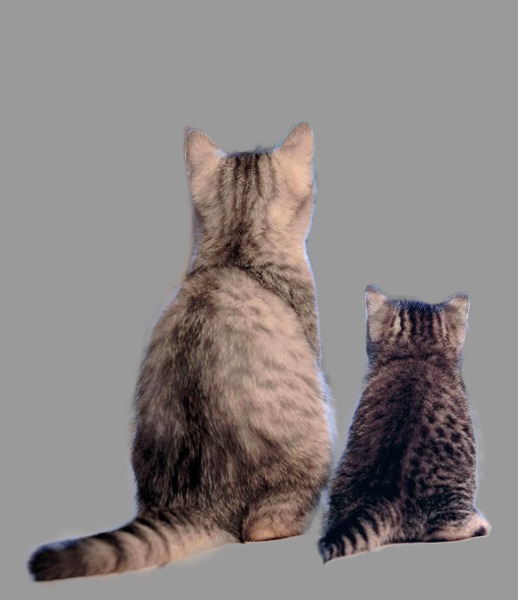}}
\subfloat[{Semantic Soft Segmentation}]{\includegraphics[height=2.7cm]{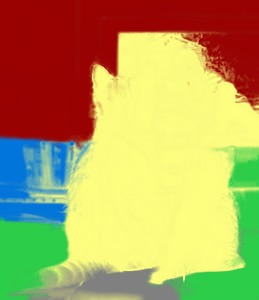}}
\subfloat[Spectral Matting]{\includegraphics[height=2.7cm]{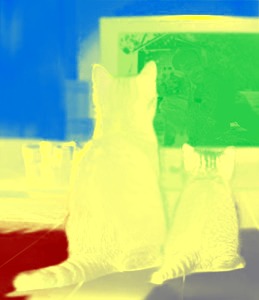}}

\subfloat{\textbf{8}\hspace{3pt}\includegraphics[height=2.7cm]{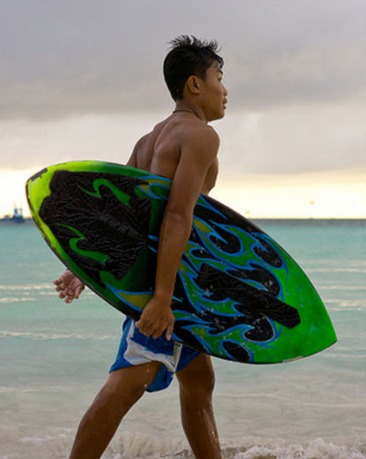}}
\subfloat{\includegraphics[height=2.7cm]{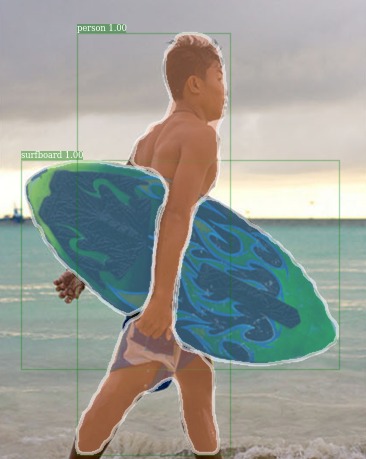}}
\subfloat{\includegraphics[height=2.7cm]{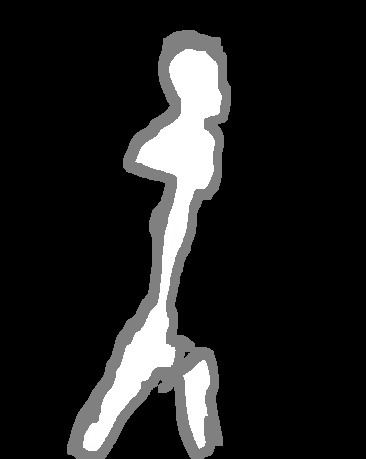}}
\subfloat{\includegraphics[height=2.7cm]{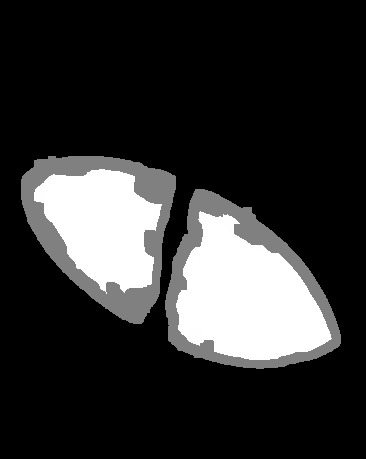}}
\subfloat{\includegraphics[height=2.7cm]{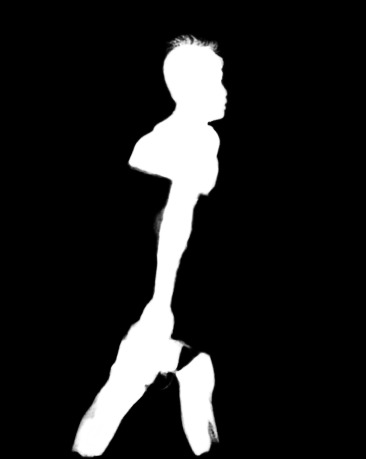}}
\subfloat{\includegraphics[height=2.7cm]{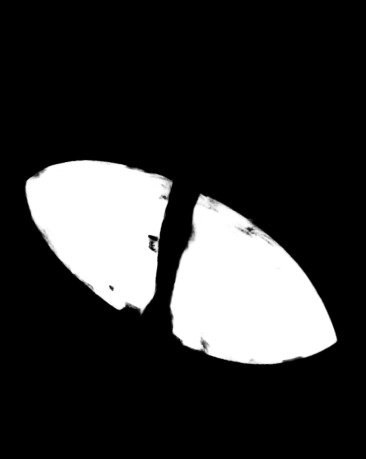}}\\
\subfloat{\includegraphics[height=2.7cm]{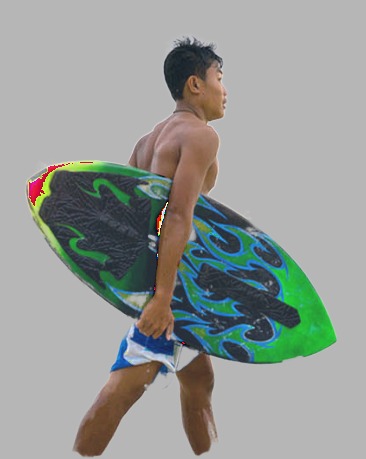}}
\subfloat{\includegraphics[height=2.7cm]{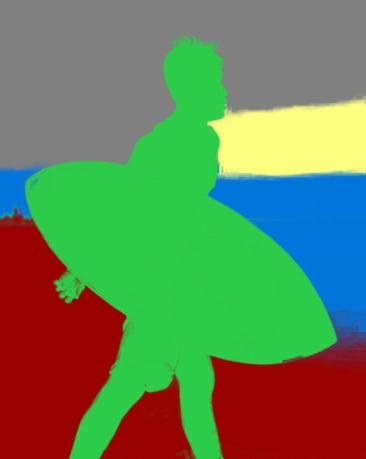}}
\subfloat{\includegraphics[height=2.7cm]{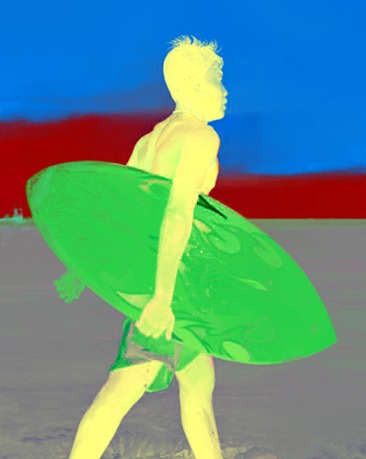}}

\subfloat{\textbf{9}\hspace{3pt}\includegraphics[height=2.7cm]{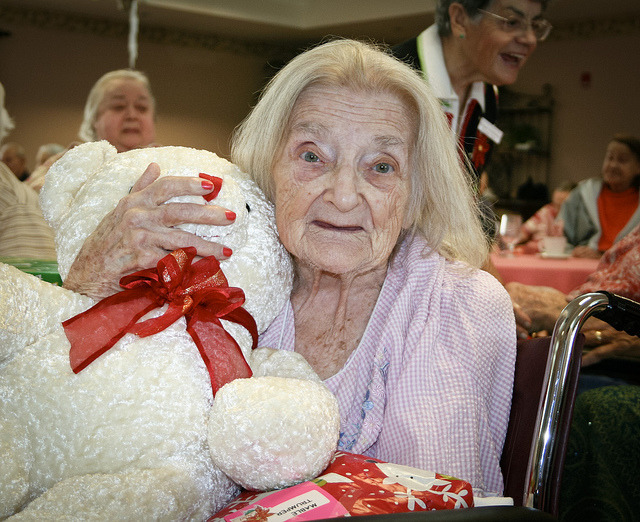}}
\subfloat{\includegraphics[height=2.7cm]{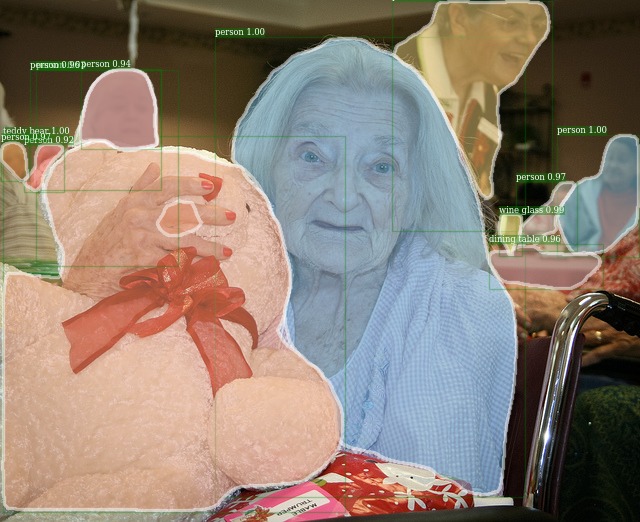}}
\subfloat{\includegraphics[height=2.7cm,trim=100 0 0 0,clip]{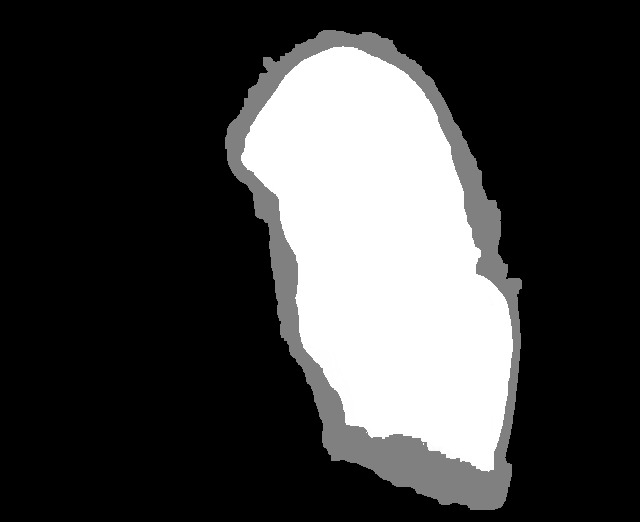}}
\subfloat{\includegraphics[height=2.7cm,trim=0 0 250 0,clip]{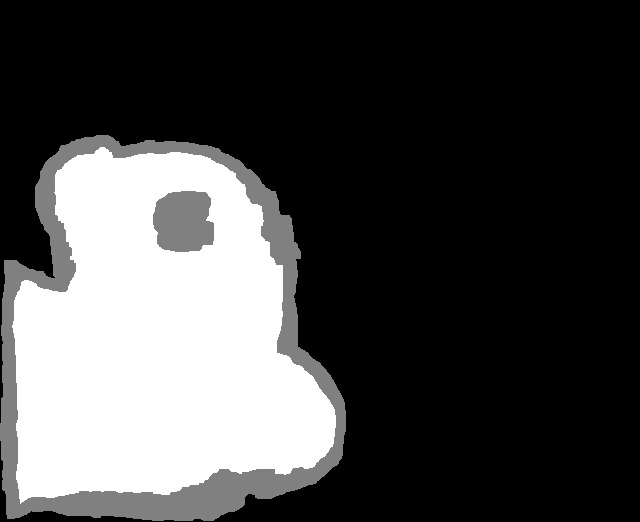}}
\subfloat{\includegraphics[height=2.7cm,trim=100 0 0 0,clip]{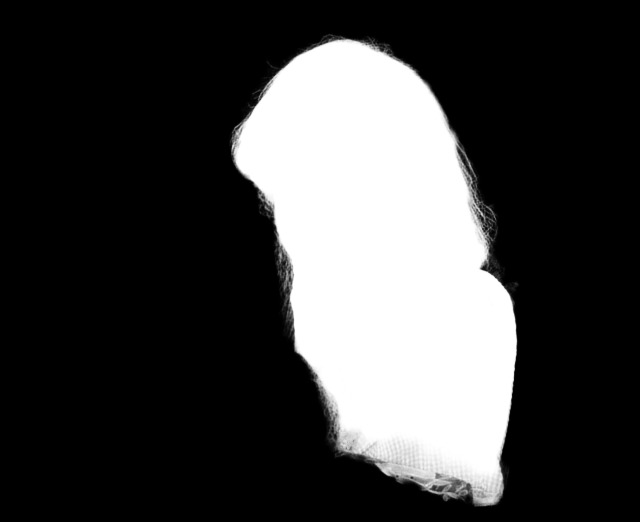}}
\subfloat{\includegraphics[height=2.7cm,trim=0 0 250 0,clip]{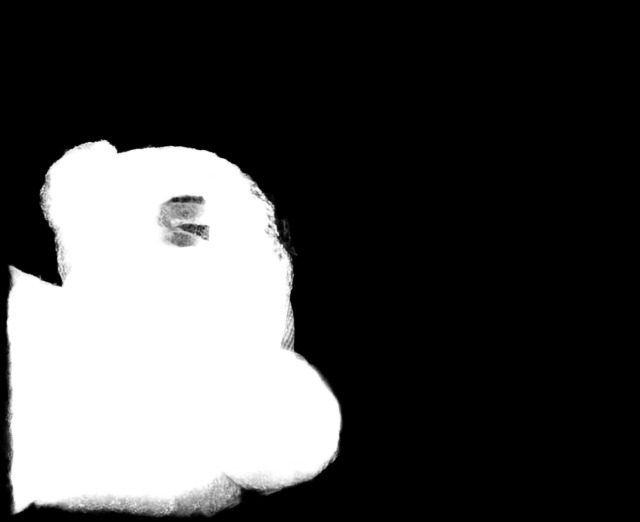}}\\
\subfloat{\includegraphics[height=2.7cm]{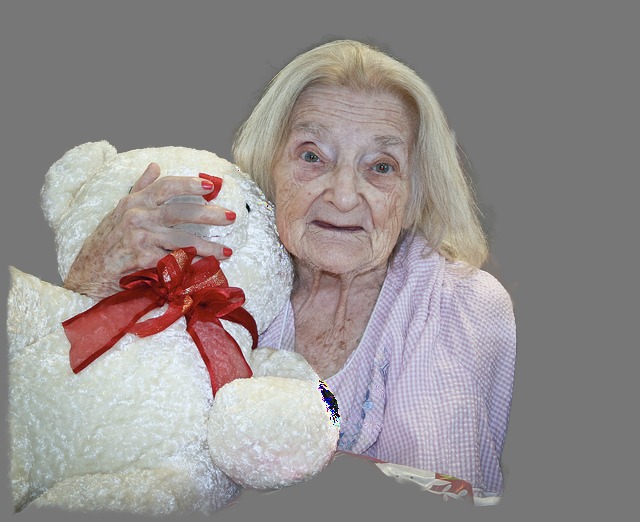}}
\subfloat{\includegraphics[height=2.7cm]{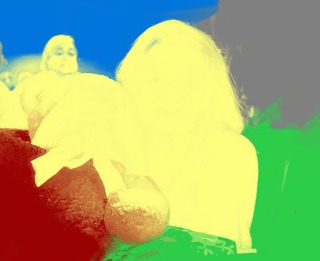}}
\subfloat{\includegraphics[height=2.7cm]{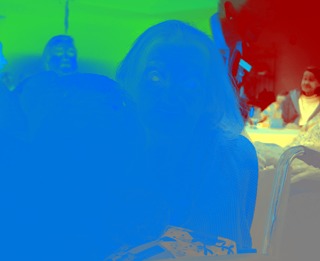}}

\caption{From left to right are RGB input image, Mask-RCNN, trimap of instance 1 at the 4th feedback loop, , trimap of instance 2 at the 4th feedback loop, alpha matte of instance 1 at the 4th feedback loop,  alpha matte of instance 2 at the 4th feedback loop, instance compositing, result of Semantic Soft Segmentation, and result of spectral matting.}
\label{resultsfig2}
\end{figure*}



\section{Conclusion}
In this work, we propose a fully automated image segmentation-matting approach in which accurate segmentation can be achieved on general image datasets such as COCO \cite{coco}. The proposed method does not require any additional user-labeled input and generates individual alpha matte for all the detected objects in the image. Our approach can be seen as both a segmentation enhancing approach and a fully automated image matting approach that works on common dataset. The performance is evaluated and compared with related approaches. Our approach performs as well as these approaches with advantage in instance-level selections. Our approach could help non-expertise in image compositing tasks and accelerate the image editing process.


\section*{Acknowledgments}

The authors would like to thank the support of Digital District Canada, especially from Dr. Jonathan Bouchard, and the scholarship support from McGill University for GH. Also thanks go to Nvidia for the donation of GPU boards for this project, and to our colleague Ibtihel Amara for her comments on the paper.



%

\end{document}